\documentclass{article}

\usepackage[final]{corl_2026}

\usepackage{times}
\usepackage[numbers]{natbib}
\usepackage{multicol}
\usepackage{wrapfig}
\usepackage{graphicx}
\usepackage{algorithm}
\usepackage{algpseudocode}
\usepackage[table]{xcolor}
\usepackage{amssymb}
\usepackage{amsmath}
\usepackage{booktabs}
\usepackage{multirow}
\usepackage{makecell}
\usepackage{subcaption}

\usepackage{hyperref}
\definecolor{techdarkblue}{rgb}{0, 0.78, 0.45}
\definecolor{forestgreen}{rgb}{0.13, 0.55, 0.13}
\definecolor{olive}{RGB}{175, 176, 3}
\definecolor{techpurple}{rgb}{0.471, 0.318, 0.663}
\definecolor{softgray}{rgb}{0.55, 0.60, 0.65}
\hypersetup{
    hidelinks,
    linkcolor=forestgreen,
    citecolor=olive,
    filecolor=techdarkblue,
    urlcolor=softgray,
    pdfview=FitH,
}
\definecolor{algo_highlight}{RGB}{38, 139, 210}
\definecolor{algo_comment}{gray}{0.5}
\definecolor{table_highlight}{rgb}{0.9, 0.9, 0.9}

\definecolor{figure_blue}{RGB}{60, 118, 175}
\definecolor{figure_orange}{RGB}{239, 134, 54}

\usepackage{cleveref}

\newcommand{\refalg}[1]{\hyperref[#1]{Alg.~\ref{#1}}}
\newcommand{\refapp}[1]{\hyperref[#1]{App.~\ref{#1}}}
\newcommand{\refeq}[1]{\hyperref[#1]{Eq.~(\ref{#1})}}
\newcommand{\reftab}[1]{\hyperref[#1]{Tab.~\ref{#1}}}

\title{Hierarchical Residual Policy Learning for Real-World Mobile Manipulation with Sparse Human Guidance}

\newcommand{\tamrealplainauthors}{Zhefei Gong, Hefeifei Jiang, Zehao Wang, Jiachen Li}
\newcommand{\tamrealauthorblock}{%
  \begin{tabular}{c}
    \\
    \fontsize{11}{13.2}\selectfont
    Zhefei Gong$^{1*}$
    \quad Hefeifei Jiang$^{2}$
    \quad Zehao Wang$^{3}$
    \quad Jiachen Li$^{3\dagger}$ \\[6pt]
    { \normalfont\fontsize{11}{13.2}\selectfont
    $^{1}$Northeastern University
    \quad
    $^{2}$University of California, Riverside
    }
    \\ { \normalfont\fontsize{11}{13.2}\selectfont
    $^{3}$Georgia Institute of Technology
    }

    \\[6pt]{\normalfont\small Project Website: \url{https://hire-moma.github.io/}}
  \end{tabular}%
}

\author{\tamrealauthorblock}

\makeatletter
\newif\ifshowrealauthors
\if@conferencefinal \showrealauthorstrue \fi
\if@preprinttype    \showrealauthorstrue \fi
\makeatother

\ifshowrealauthors
  \hypersetup{pdfauthor={\tamrealplainauthors}}
\else
  \hypersetup{pdfauthor={Anonymous Submission}}
\fi

\hypersetup{pdftitle={Hierarchical Residual Policy Learning for Real-World Mobile Manipulation with Sparse Human Guidance}}

\begin{document}
\maketitle

\ifshowrealauthors
  \begingroup
    \renewcommand{\thefootnote}{\fnsymbol{footnote}}
    \footnotetext[1]{Work done during an internship at the Trustworthy Autonomous Systems Laboratory (TASL)}
    \footnotetext[2]{Corresponding author: \href{mailto:jiachen_li@gatech.edu}{jiachen\_li@gatech.edu}}
  \endgroup
\fi

\begin{abstract}
Mobile Manipulation (MoMa) is a cornerstone of autonomous robotic systems, yet real-world fine-tuning remains a formidable challenge because of high-dimensional action spaces and intricate coupling between heterogeneous modules.
In this paper, we propose \textbf{HiRe-MoMa} (\textbf{\underline{Hi}}erarchical \textbf{\underline{Re}}sidual \textbf{\underline{Mo}}bile \textbf{\underline{Ma}}nipulation), a systematic framework for stable and sample-efficient real-world fine-tuning of complex MoMa systems.
At its core, HiRe-MoMa adopts a Hierarchical Residual Architecture that fine-tunes a frozen base policy.
A Policy-Agnostic Residual learns additive corrections atop the base policy, while Hierarchical Decoupling splits the residual into mobile-base and manipulator-arm heads with unidirectional conditioning, mitigating gradient interference and aligning corrections with the platform's physical hierarchy.
To improve sample efficiency in data-expensive real-world settings, we propose Expert-Guided Value Alignment, which leverages sparse human interventions with under 20\% training coverage
to reshape the critic’s value landscape, effectively guiding autonomous exploration through expert priors.
Extensive real-world experiments on the TIAGo++ platform demonstrate that HiRe-MoMa consistently enhances task performance with only around 30 minutes of online interaction, regardless of the base policy architecture.
To our knowledge, this is among the first frameworks to enable efficient on-robot fine-tuning for mobile manipulation in the real world, providing a practical paradigm for deploying complex robotic systems in dynamic, human-centric environments.
Further details can be found on our \href{https://hire-moma.github.io/}{project website}.

\end{abstract}


\section{Introduction}
\label{sec:introduction}

Mobile manipulation is a key step toward multi-purpose robotic systems that operate in human-centric environments.
Performing long-horizon and multi-object tasks, from domestic chores to industrial logistics, requires the coordinated operation of heterogeneous modules such as mobile bases, robotic arms, and sensory heads.
Despite its potential, autonomous mobile manipulation remains challenging due to high-dimensional action spaces, tightly coupled inter-module dynamics, and the non-stationarity of real-world deployment.
Recent advancements have leveraged Vision-Language Models (VLMs) for semantic reasoning~\cite{shah2025bumble,jiang2026navtrust,zhang2026commcp,yan2025rdd,chakraborty2026toward}, generative modeling for robust imitation learning~\cite{jiang2025brs,ghosh2025robust}, and simulation-to-real reinforcement learning for skill acquisition~\cite{Hu-RSS-23}.
However, a single-shot trained policy often fails to generalize to the stochasticity of diverse real-world deployment scenarios.
Post-training or fine-tuning is thus essential to bridge the performance gap. While fine-tuning has seen success in isolated manipulation or locomotion tasks~\cite{amin2025pi, smith2022legged}, its application to the coupled, high-dimensional domain of mobile manipulation remains largely unexplored.

\begin{figure}[t]
   \centering
   \includegraphics[width=1.00\linewidth]{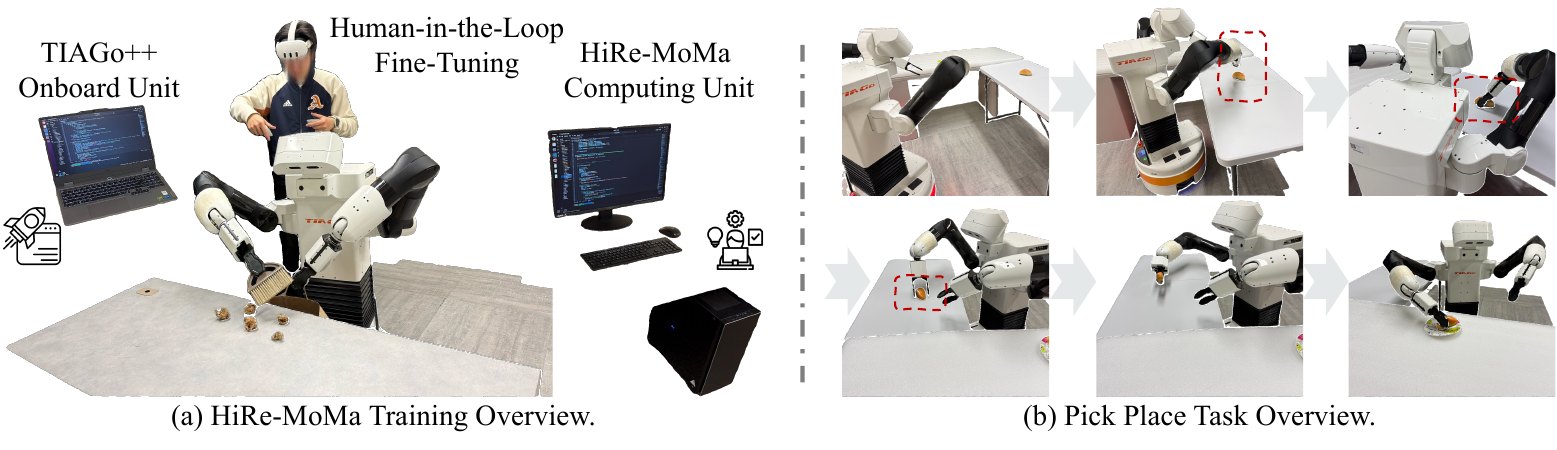}
   \caption{\small{
   \textbf{HiRe-MoMa System Overview.}
   \textbf{(a) Training Paradigm:}
   A real-world dual-unit configuration consisting of an onboard controller for real-time interaction and a computing unit for policy optimization and action prediction.
   \textbf{(b) Coordinated Execution:}
   The red dashed lines illustrate the stringent spatiotemporal precision required for base-arm synchronization.
   In mobile manipulation, marginal tracking deviations are often magnified by kinematic coupling, causing even routine grasping maneuvers to fail.
   }}
   \label{fig:teaser}
\end{figure}

Real-world fine-tuning, however, is shaped by a series of fundamental trade-offs.
Imitation learning (IL), particularly DAgger-based approaches~\cite{ross2011reduction}, offers a straightforward optimization objective but is constrained by inherent expert data scarcity and a lack of autonomous exploration, which often leads to limited generalization on out-of-distribution states and compounding errors at deployment.
Reinforcement learning (RL) enables broader state-space exploration~\cite{liu2025what} but faces significant obstacles when paired with modern generative base policies, whose intractable log-likelihoods preclude standard policy gradient methods~\cite{zhang2025reinflow}.
Residual RL offers a promising and practical middle ground by learning additive corrections atop frozen base policies for model-agnostic adaptation~\cite{jiang2025transic, ankile2025imitation}, yet extending it to mobile manipulation introduces a distinct challenge: \textit{inter-module coupling}.
As shown in~\autoref{fig:teaser}, a task as routine as grasping becomes difficult in the MoMa context: it requires tight spatiotemporal coordination between the mobile base and the arm, and any misalignment in the base's approach rapidly propagates as a sizable pose error at the manipulator.
Without explicit structural priors, the high-dimensional action space and non-stationary inter-module dynamics often hinder convergence and ultimately prevent the precision required for physical interaction.

In this paper, we propose \textbf{HiRe-MoMa} (\textbf{\underline{Hi}}erarchical \textbf{\underline{Re}}sidual \textbf{\underline{Mo}}bile \textbf{\underline{Ma}}nipulation), a framework for real-world fine-tuning of mobile manipulation systems, built around three components.
First, our \textbf{Policy-Agnostic Residual Learning} paradigm adds a learnable residual atop a \emph{frozen} base policy, sidestepping the intractable log-likelihoods of generative architectures and allowing the same procedure to refine diverse pre-trained policies.
Second, to address this base-arm coupling specific to mobile manipulation, we introduce a \textbf{Hierarchical Decoupling} structure with unidirectional conditioning: the manipulator's residual head is conditioned on the base's correction, reflecting the system's physical hierarchy.
Third, an \textbf{Expert-Guided Value Alignment} loss leverages sparse human interventions to reshape the critic's value landscape toward expert-preferred behaviors, improving sample efficiency without overriding RL exploration.
HiRe-MoMa is realized as an asynchronous actor-learner system on a single robot:
an onboard rollout worker executes the policy
while a central learner optimizes the residual on a workstation,
enabling on-robot fine-tuning in the real world.

Overall, our main contributions are as follows:\\
\textbf{1) Systematic Fine-tuning Framework}: We present an asynchronous, model-agnostic framework for the real-world fine-tuning of mobile manipulation systems, enabling efficient adaptation in this high-dimensional and coupled robotic domain with 25--40 minutes of online interaction.
\textbf{2) Hierarchical Residual Architecture}: We introduce independent residual policies for distinct modules, coupled with a unidirectional conditioning mechanism to ensure stable and logically coherent coordinated control.
\textbf{3) Expert-Guided Alignment}: We develop an auxiliary human intervention alignment loss that leverages sparse expert guidance ($\leq$20\% of collected data) to improve sample efficiency and accelerate convergence.
\textbf{4) Empirical Validation on Diverse MoMa Tasks}: We validate HiRe-MoMa on three real-world TIAGo++ tasks of increasing complexity, demonstrating consistent improvements across two distinct base policy architectures (Diffusion Policy and ACT).

\section{Related Work}
\label{sec:relatedwork}

\begin{figure*}[!t]
   \centering
   \includegraphics[width=1.0\linewidth]{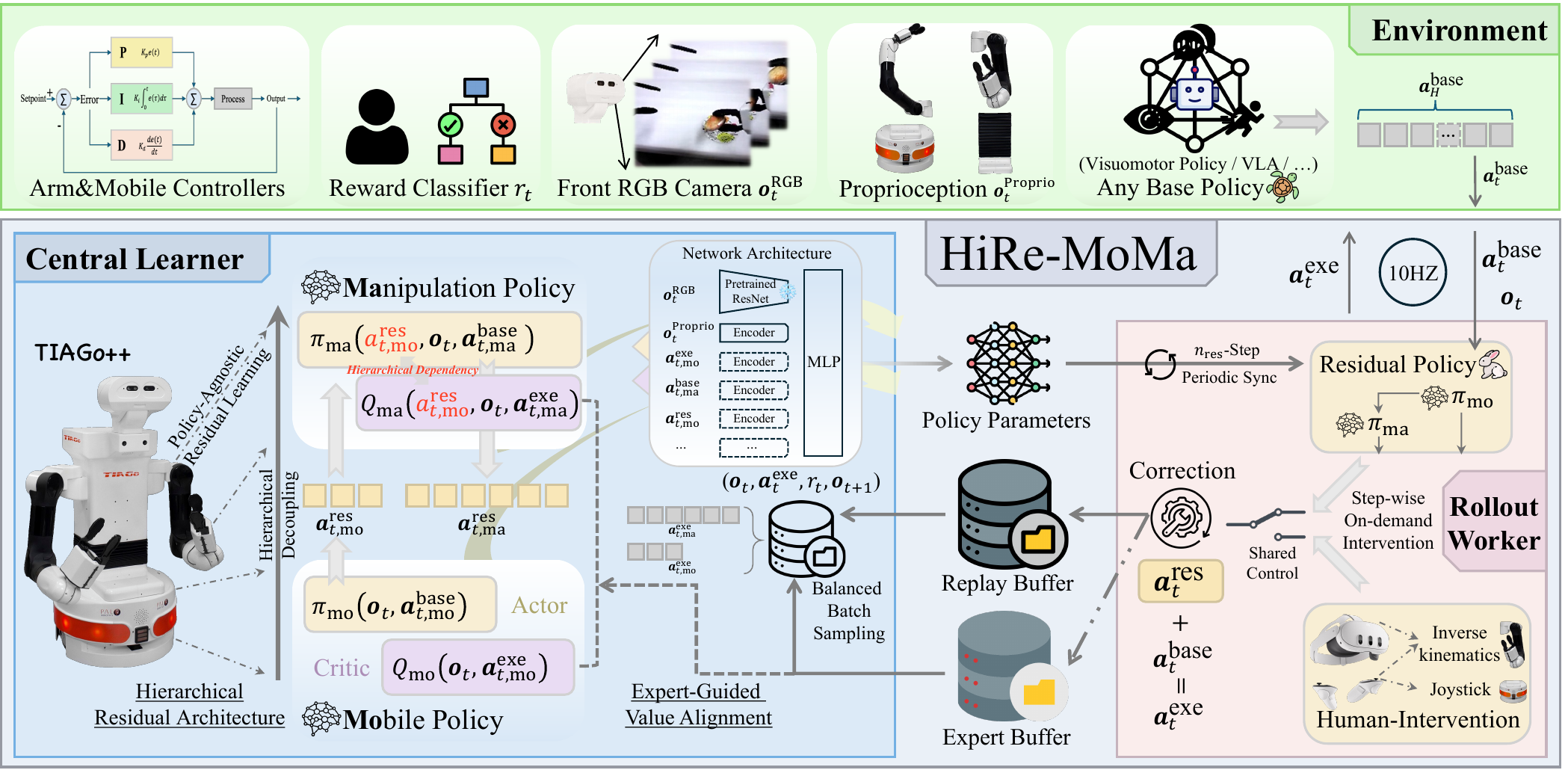}
   \caption{\small{
   \textbf{HiRe-MoMa Framework}
   enables policy-agnostic real-world fine-tuning of mobile manipulators through an asynchronous residual reinforcement learning paradigm.
   \textbf{(Right) Shared-Control Rollout:}
   an onboard rollout worker augments a frozen base policy with learnable residuals;
   sparse human interventions during critical or hazardous phases are stored in an expert buffer.
   \textbf{(Left) Asynchronous Central Learner:}
   the residual policy is optimized along two pillars: a) a \textit{Hierarchical Residual Architecture} (\autoref{sec:method_hierarchical_residual_architecture}) that decouples the residual into mobile and manipulator heads with unidirectional conditioning, embedding the platform's physical hierarchy directly into the policy; and b) an \textit{Expert-Guided Value Alignment} loss (\autoref{sec:method_hil_auxiliary_loss}) that fuses sparse environment rewards with sparse human corrections to reshape the critic toward expert-preferred behaviors.
   }}
   \vspace{-0.3cm}
   \label{fig:frame}
\end{figure*}

\vspace{-1mm}

\textbf{Real-World Reinforcement Learning}
offers a principled path to bridging the sim-to-real gap~\cite{XuC1-RSS-25}, with sample efficiency as the central challenge.
The SERL framework~\cite{luo2024serl} demonstrated the efficacy of off-policy RL for robot fine-tuning, inspiring subsequent residual correction methods~\cite{chen2025conrft, ankile2025residual}.
Existing residual and HIL systems~\cite{luo2025precise, chen2025conrft, jiang2025transic, ankile2025imitation, ankile2025residual} fine-tune fixed-base arms, where base--arm gradient interference does not arise;
HiRe-MoMa instead targets policy-agnostic, on-robot fine-tuning of a whole-body mobile manipulator,
without simulation-pre-trained latent action spaces~\cite{hu2025slac}.

\textbf{Policy Post-Training}
of base policies~\cite{chi2025diffusion, zhao2025vlas} via RL is hampered by intractable log-likelihoods, motivating surrogate objectives~\cite{zhang2025reinflow, chen2025pi_, wang2024diffusion, pmlr-v267-ma25d} or policy-agnostic RL~\cite{mark2024policy, chen2025conrft}. Residual RL offers a practical middle ground by learning additive corrections atop frozen policies~\cite{ankile2025residual, gong2025robust,Wang_2026_CVPR,11696068,gao2026guides}, with human-in-the-loop variants~\cite{kelly2019hg, hu2025rac,kim2025human} further mitigating distributional drift.
Yet simple residuals cannot address the cross-module coupling inherent in MoMa:
the spatiotemporal dependency between base and arm demands structured decoupling that flat residual methods do not provide.

\textbf{Mobile Manipulation}
poses unique challenges owing to its high-dimensional action space and tight base-arm coupling~\cite{jiang2025brs, chen2025acdit}. Most RL methods still rely on simulation-pre-trained priors~\cite{Hu-RSS-23, zhang2025catch} or expensive skill-discovery phases~\cite{hu2025slac, Mendonca2024Continuously} to bridge the sim-to-real gap. HiRe-MoMa instead operates directly in the raw action space of any pre-trained policy, modeling hierarchical inter-module dependencies under a sparse success-only reward without manual reward shaping.

\textbf{Human-in-the-Loop}
corrections serve as safety buffers~\cite{luo2025precise, amin2025pi}, imitation data sources~\cite{Liu-RSS-23}, or auxiliary supervision signals~\cite{li2022efficient, wang2025inference, korkmaz2025mile} in robot RL. Beyond these roles, interventions implicitly encode expert value preferences~\cite{amin2025pi}; HiRe-MoMa exploits this by treating corrections as functional constraints on the critic’s value surface, directly reshaping the value landscape to reflect expert priorities.
A broader discussion of related work on these four topics is provided in~\refapp{appendix:related_work}.

\vspace{-0.1cm}
\section{Problem Formulation}
\vspace{-0.1cm}
\label{sec:preliminary}

We study the problem of fine-tuning pre-trained base policies $\pi^{\text{base}}$~\cite{pmlr-v305-black25a, chi2025diffusion, gong2025carp} for complex robotic tasks in unstructured real-world environments.
While Supervised Fine-Tuning (SFT) via imitation learning offers a stable convergence path, its efficacy is fundamentally bounded by the coverage and quality of pre-collected demonstrations, which are typically scarce and sub-optimal in practice.
Conversely, Reinforcement Learning (RL) enables the agent to extend beyond the constraints of static datasets through active exploration, fostering improved robustness and task proficiency.
This online refinement is especially important for high-dimensional systems such as mobile manipulators, where conventional zero-shot transfer leaves a persistent deployment-time gap.
We formalize the fine-tuning process as a Partially Observable Markov Decision Process (POMDP), defined by the tuple $\mathcal{M} = (\mathcal{S}, \mathcal{A}, \mathcal{O}, \mathcal{P}, R, \gamma)$.
$\mathcal{S}$ represents the underlying hidden environment states, and $\mathcal{A} \subset \mathbb{R}^d$ denotes the high-dimensional continuous action space (e.g., joint velocities or Cartesian poses).
The observation space $\mathcal{O}$ comprises multi-modal sensory inputs, including RGB images and robot proprioception, while the transition dynamics $\mathcal{P}(s'|s, a)$ are governed by real-world physical interactions.
To minimize reward engineering and remain task-agnostic, we adopt a sparse binary reward function $R(s, a) \in \{0, 1\}$, where $1$ denotes task success and $0$ otherwise.
The objective is to refine the policy $\pi$ to maximize the expected cumulative discounted return:
$J(\pi) = \mathbb{E}_{\tau \sim \pi} \left[ \sum_{t=0}^{\infty} \gamma^t R(s_t, a_t) \right]$,
where $\gamma \in (0, 1]$ is the discount factor.
This formulation drives policy refinement toward higher task success rates and improved temporal efficiency under real-world execution.

\section{Method}
\label{sec:method}

To facilitate efficient fine-tuning for high-dimensional mobile manipulation, we propose \textbf{HiRe-MoMa} (\textbf{\underline{Hi}}erarchical \textbf{\underline{Re}}sidual \textbf{\underline{Mo}}bile \textbf{\underline{Ma}}nipulation), a model-agnostic framework designed for stable real-world adaptation via human-robot interaction (\autoref{fig:frame}). The framework consists of three components:
\textbf{(i)} a \emph{policy-agnostic residual} that learns additive corrections atop a frozen prior $\pi^{\text{base}}$ without structural assumptions on the base policy;
\textbf{(ii)} a \emph{hierarchical decoupling} mechanism that mitigates optimization interference between heterogeneous modules (e.g., mobile base and manipulator arm) by enforcing a unidirectional dependency aligned with the system's physical hierarchy;
\textbf{(iii)} an \emph{expert-guided value alignment} loss that leverages sparse human interventions to reshape the critic's value landscape, improving sample efficiency under safe real-world exploration.
We detail each component below; full implementation details are provided in~\refapp{appendix:impl_details}.

\subsection{Hierarchical Residual Architecture}
\label{sec:method_hierarchical_residual_architecture}

\textbf{Policy-Agnostic Residual Learning.}
\label{sec:method_architecture_residual}
Fine-tuning pre-trained policies via
reinforcement learning
often entails restrictive structural assumptions, particularly when involving intractable log-likelihood estimations~\cite{zhang2025reinflow,wang2024diffusion}.
To achieve structural agnosticism across arbitrary base policies $\pi^{\text{base}}$, we introduce an auxiliary residual policy $\pi_{\theta}^{\text{res}}$ that learns to predict additive corrections atop a frozen prior.
Given the high sample complexity of real-world interactions, we adopt an off-policy, value-based paradigm~\cite{luo2025precise} utilizing Temporal Difference (TD) learning~\cite{sutton1988learning}.
To stabilize value estimation and mitigate erratic exploration, $Q_\phi$ is optimized via a conservative target derived from the minimum of a critic ensemble.
\begin{equation}
\begin{aligned}
    Q_{\phi}(\mathbf{o}_t, \mathbf{a}_t) \leftarrow r_t + \gamma \mathbb{E}_{\mathbf{o}_{t+1} \sim \mathcal{P}} \left[ Q_{\phi'}(\mathbf{o}_{t+1}, \mathbf{a}_{t+1}^{\text{exe}}) \right], \quad
    \text{where} \quad \mathbf{a}_{t+1}^{\text{exe}} = \mathbf{a}_{t+1}^{\text{base}} + \pi_{\theta}^{\text{res}}(\mathbf{o}_{t+1}, \mathbf{a}_{t+1}^{\text{base}}).
\end{aligned}
\label{eq:residual_learn_critic}
\end{equation}
The residual policy $\pi^{\text{res}}_{\theta}$ is optimized to maximize the expected return while remaining localized around the base policy's trajectory via entropy regularization $\mathcal{H}$~\cite{haarnoja2018soft}:
\begin{equation}
\begin{aligned}
    \mathbb{E}_{\pi^{\text{res}}_\theta} [Q_\phi(\mathbf{o}_t, \mathbf{a}^{\text{exe}}_{t})] + \alpha \mathcal{H}(\pi^{\text{res}}_\theta(\cdot|\mathbf{o}_{t})), \quad
    \text{where} \quad \mathbf{a}_{t}^{\text{exe}} = \mathbf{a}_{t}^{\text{base}} + \pi_{\theta}^{\text{res}}(\mathbf{o}_{t}, \mathbf{a}_{t}^{\text{base}}).
\end{aligned}
\label{eq:residual_learn_policy}
\end{equation}
where $\alpha$ is an adaptive temperature parameter.
To stabilize exploration and prevent corrections from exceeding the residual module's capacity, we constrain $\| \mathbf{a}^{\text{res}} \| \le \epsilon$ with $\epsilon \in (0,1]$.
The resulting residual policy $\pi^{\text{res}}$ remains agnostic to $\pi^{\text{base}}$'s architecture, providing flexibility across heterogeneous robotic platforms (\refalg{algo:hire_moma}, Rollout Worker and Critic Update blocks).
This module also serves as the building block for the hierarchical decoupling introduced next.

\textbf{Hierarchical Decoupling.}
\label{sec:method_architecture_hierarchical}
Mobile manipulation requires high-dimensional coordination across heterogeneous modules (mobile base, manipulator arm, torso).
Treating the joint action space as a monolithic output couples the gradients of these modules, which leads to interference and unstable optimization.
Mobile manipulation tasks naturally exhibit a unidirectional inter-module dependency: the manipulator's reaching motion is conditioned on the base's positioning rather than the reverse.
Such structural priors have been shown to align fine-tuned corrections with the system's physical hierarchy~\cite{jiang2025brs, chen2025acdit}.
Leveraging our modular residual design, we first decouple the action space into independent components, denoted as $\mathbf{a}^{\text{res}} = [\mathbf{a}_{\text{mo}}^{\text{res}}, \mathbf{a}_{\text{ma}}^{\text{res}}]$ for the mobile base and manipulator, respectively.
This decomposition mitigates the gradient interference inherent in high-dimensional RL by reducing the cross-module optimization conflict, which is critical for training stability.
We then encode this dependency through unidirectional conditioning at each timestep $t$:
\begin{equation}
    \mathbf{a}_{t, \text{mo}}^{\text{res}} = \pi^{\text{res}}_{\theta_{\text{mo}}}(\mathbf{o}_t, \mathbf{a}_{t, \text{mo}}^{\text{base}})
    \rightarrow
    \mathbf{a}_{t, \text{ma}}^{\text{res}} = \pi^{\text{res}}_{\theta_{\text{ma}}}(\mathbf{a}_{t, \text{mo}}^{\text{res}}, \mathbf{o}_t, \mathbf{a}_{t, \text{ma}}^{\text{base}}).
\label{eq:hierarchy_condition}
\end{equation}
By treating the mobile correction $\mathbf{a}_{t, \text{mo}}^{\text{res}}$ as an upstream conditioning context for the manipulator policy $\pi^{\text{res}}_{\theta_{\text{ma}}}$ and its corresponding critic $Q_{\phi_{\text{ma}}}$, we enforce a unidirectional structural constraint that mirrors the physical coupling of the system.
This design mitigates the non-stationarity of joint MoMa training by aligning arm refinements with the platform's evolving coordinate frame, yielding more coherent coordination and stable convergence (\refalg{algo:hire_moma_hierarchy}, \textcolor{algo_highlight}{highlighted in light blue}; full pseudocode in~\refapp{appendix:pseudocode}).
The conditioning chain naturally extends to deeper hierarchies, accommodating additional degrees of freedom (DoF) or specialized modules as task complexity grows.

\subsection{Expert-Guided Value Alignment}
\label{sec:method_hil_auxiliary_loss}
The high dimensionality and exploration cost of real-world mobile manipulation pose significant challenges for vanilla RL, even with constrained residual spaces.
To accelerate convergence and enhance safety, we integrate human interaction into the training loop via a shared-control paradigm.
Human experts can intervene during critical or high-risk task stages to provide corrective guidance, which is stored in a dedicated intervention buffer.
During training, we perform balanced sampling~\cite{luo2025precise} from both the autonomous exploration buffer and the human intervention buffer to align the agent’s learning with expert priors.
Rather than direct action imitation (e.g., Behavior Cloning), which is sensitive to teleoperation noise, we use human corrections to reshape the value landscape.
This corresponds to the \textit{Expert-Guided Critic Alignment} block in~\refalg{algo:hire_moma} (lines 28--29, \textcolor{algo_highlight}{highlighted in light blue}): an auxiliary supervised objective on $Q_{\phi}$ that provides a stable, conservative target~\cite{li2022efficient}.
The objective minimizes the following preference loss:
\begin{equation}
    \lambda_{\text{imi}} \mathbb{E}_{\mathbf{o}_t, \mathbf{a}_t^{\text{res}}, \mathbf{a}_t^{\text{h}}} \left[ \log \left( 1 + \exp \left( Q_{\phi}(\mathbf{o}_t, \mathbf{a}_t^{\text{res}}) - Q_{\phi}(\mathbf{o}_t, \mathbf{a}_t^{\text{h}}) \right) \right) \right],
    \label{eq:hil_imitation_loss}
\end{equation}
where $\mathbf{a}_t^{\text{h}}$ represents the expert correction from the human-intervened policy $\pi^{\text{res,h}}$, and $\lambda_{\text{imi}}$ scales the auxiliary supervision.
This formulation encourages the critic to be optimistic toward human actions while remaining pessimistic toward the agent's autonomous exploration.
By embedding human preferences directly into the proxy value function, the agent aligns with expert priors more efficiently, leading to faster and more stable convergence in complex environments.

\section{Experiments}
\label{sec:experiments}

We evaluate HiRe-MoMa through real-world experiments on the TIAGo++ mobile manipulator, designed to isolate the contribution of each method component and to test adaptability across base-policy architectures and long-horizon tasks. Our study addresses the following Research Questions: \\
\textbf{[RQ1]} \textbf{Performance \& Generality}: Does HiRe-MoMa outperform existing residual refinement baselines across diverse architectures in a model-agnostic manner?\\
\textbf{[RQ2]} \textbf{Hierarchical Architecture}: Does Hierarchical Decoupling (splitting base/arm heads with unidirectional conditioning) improve success rate and reduce episode length on MoMa tasks?\\
\textbf{[RQ3]} \textbf{Value Alignment}: Does the Expert-Guided Value Alignment loss further improve sample efficiency and final performance on top of Hierarchical Decoupling?\\
\textbf{[RQ4]} \textbf{Interpretability}: Can the predicted residual corrections be qualitatively validated through visualization to ensure they provide physically meaningful refinements?

\begin{table}[!t]
    \centering
    \caption{\small{
    \textbf{Comparison with Fine-Tuning Baselines.}
    All methods use a frozen DP as the base policy.
    \textit{+HG-DAgger}~\cite{kelly2019hg} overfits to sparse human corrections, and \textit{+Flat Residual} (equivalent to HIL-SERL~\cite{luo2025precise} Residual) is limited by inter-module gradient interference in its monolithic action space.
    The full HiRe-MoMa consistently outperforms both in success rate and execution efficiency.
    Metrics are averaged over 20 trials per task.
    }}
    \resizebox{\columnwidth}{!}{
    \begin{tabular}{r|c|cccc|cccc}
    &\multirow{2}{*}{\makecell[c]{\textbf{Training}\\\textbf{Time}\\\textbf{(mins)}}}
    &\multicolumn{4}{c}{\textbf{Success Rate (\%)}}
    &\multicolumn{4}{c}{\textbf{Episode length (steps)}}
    \\
    \textbf{Task} &
    &\makecell[c]{Base\\(DP)}
    &\makecell[c]{+HG-DAgger\\\cite{kelly2019hg}}
    &\makecell[c]{+Flat Residual\\\cite{luo2025precise}}
    &\makecell[c]{HiRe-MoMa\\(Full)}
    &\makecell[c]{Base\\(DP)}
    &\makecell[c]{+HG-DAgger\\\cite{kelly2019hg}}
    &\makecell[c]{+Flat Residual\\\cite{luo2025precise}}
    &\makecell[c]{HiRe-MoMa\\(Full)}
    \\
    \hline
    Place Only   &25 &50.0 &10.0 (-80\%)  &60.0 (+20\%)  &\textbf{95.0} (+90\%) &342 &420 (0.81x) &329 (1.03x) &\textbf{235} (1.46x) \\
    Sweep Litter &30 &45.0 &5.0 (-89\%)  &40.0 (-11\%)   &\textbf{90.0} (+100\%) &359 &478 (0.75x) &394 (0.91x) &\textbf{271} (1.32x) \\
    Pick Place   &40 &22.5 &5.0 (-78\%) &27.5 (+22\%)   &\textbf{40.0} (+78\%) &664 &742 (0.89x) &649 (1.02x) &\textbf{552} (1.20x) \\
    \hline
    \rowcolor{table_highlight}
    \textbf{Average}  &31.7 &39.2 &6.7 (-83\%) &42.5 (+8\%) &\textbf{75.0} (+91\%) &455.0 &546.7 (0.82x) & 457.3 (0.99x) &\textbf{352.7} (1.29x) \\
    \end{tabular}
    }
    \label{tab:exp_main_results}
\end{table}

\subsection{Experimental Setup}
\label{sec:exp_setup}
\textbf{Platform.}
We evaluate HiRe-MoMa on the TIAGo++ mobile manipulator with the torso and head fixed,
leaving a 17-D action space (3-D base velocity; per arm, a 6-D end-effector delta pose and a 1-D gripper),
of which each task uses a subset.
Human interventions are delivered through a Meta Quest 3 teleoperation interface~\cite{dass2024telemoma}
with absolute priority over the residual policy;
for every method, interventions are collected online by the same operator under an identical,
method-blind protocol.
Following HIL-SERL~\cite{luo2025precise}, rollout and learning run asynchronously
(\autoref{fig:teaser}; more details in~\refapp{appendix:exp_setup}).

\begin{table}[!t]
    \centering
    \caption{\small{
    \textbf{Build-Up Ablation of HiRe-MoMa Components.}
    Each column adds one method component on top of the previous:
    \textit{+Flat Residual} is a single residual head over the joint action space (equivalent to HIL-SERL~\cite{luo2025precise} Residual);
    \textit{+Hierarchical Decoupling} splits the residual into base/arm heads with unidirectional conditioning (\autoref{sec:method_hierarchical_residual_architecture});
    \textit{+Alignment} is the full HiRe-MoMa with Expert-Guided Value Alignment (\autoref{sec:method_hil_auxiliary_loss}).
    Results show base policy $\rightarrow$ fine-tuned performance on tasks \textit{Place Only} and \textit{Sweep Litter}.
    }}
    \resizebox{\columnwidth}{!}{
    \begin{tabular}{r|c|ccc|ccc}
    &\multirow{2}{*}{\makecell[c]{\textbf{Training}\\\textbf{Time}\\\textbf{(mins)}}}
    &\multicolumn{3}{c}{\textbf{Success Rate (\%)}}
    &\multicolumn{3}{c}{\textbf{Episode length (steps)}}
    \\
    \textbf{Task} &
    &\makecell[c]{+Flat\\Residual}
    &\makecell[c]{+Hierarchical\\Decoupling}
    &\makecell[c]{+Alignment\\(Full)}
    &\makecell[c]{+Flat\\Residual}
    &\makecell[c]{+Hierarchical\\Decoupling}
    &\makecell[c]{+Alignment\\(Full)}
    \\
    \hline
    Place Only   &25 &50$\rightarrow$60 &50$\rightarrow$85 &50$\rightarrow$\textbf{95} &342$\rightarrow$329 &342$\rightarrow$283 &342$\rightarrow$\textbf{235} \\
    Sweep Litter &30 &45$\rightarrow$40 &45$\rightarrow$85 &45$\rightarrow$\textbf{90} &359$\rightarrow$394 &359$\rightarrow$304 &359$\rightarrow$\textbf{271} \\
    \hline
    \rowcolor{table_highlight}
    \textbf{Average}  &27.5 &47.5$\rightarrow$50.0 &47.5$\rightarrow$85.0 &47.5$\rightarrow$\textbf{92.5} &350.5$\rightarrow$361.5 &350.5$\rightarrow$293.5 &350.5$\rightarrow$\textbf{253.0} \\
    \end{tabular}}
    \label{tab:exp_ablation_buildup}
\end{table}

\begin{figure*}[!t]
    \centering
    \includegraphics[width=0.95\linewidth]{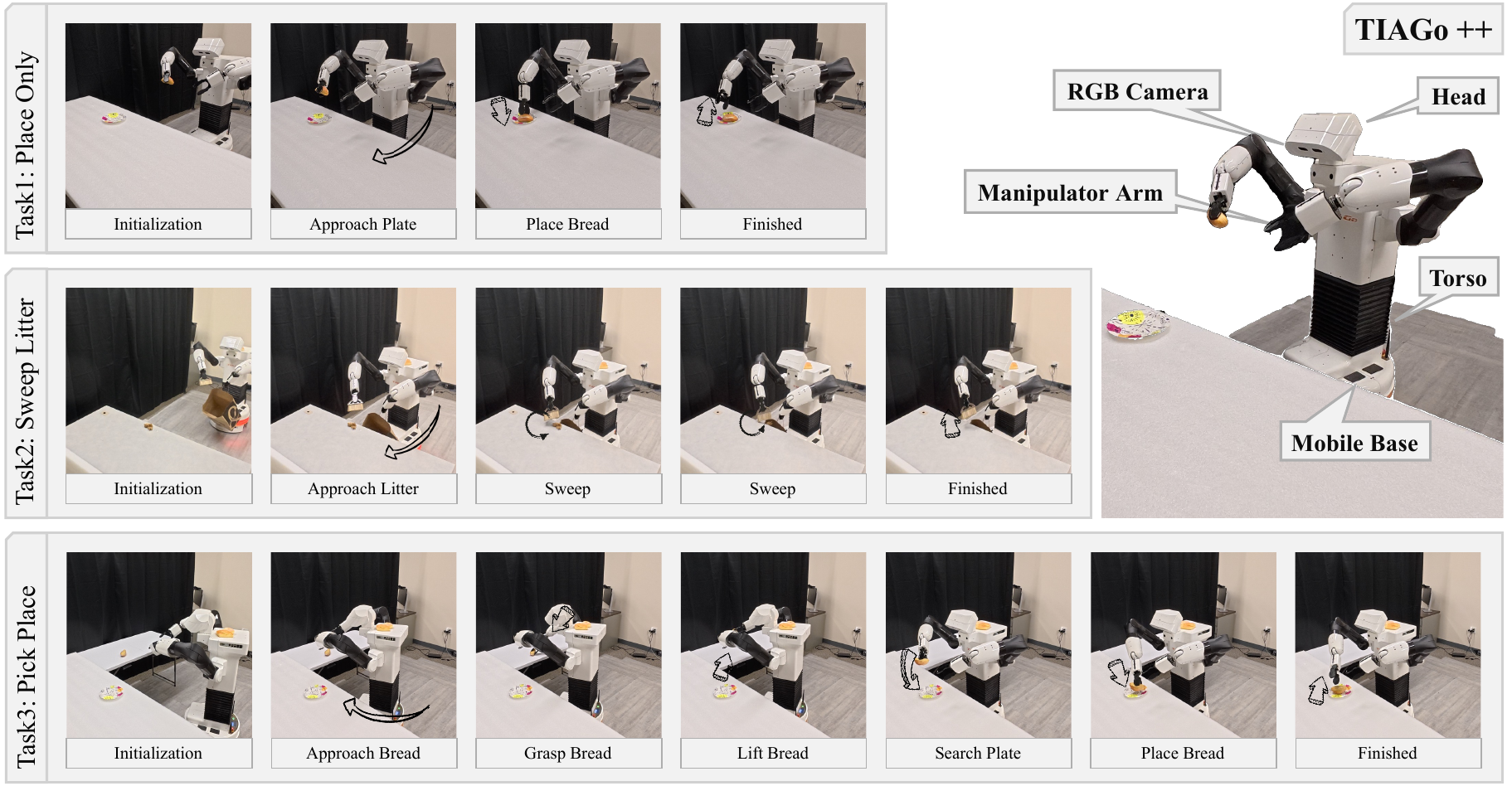}
    \caption{\small{
    \textbf{Real-World Experimental Tasks.}
    Our evaluation encompasses three
    tasks, \textit{Place Only}, \textit{Sweep Litter}, and \textit{Pick Place}, that demand varying levels of inter-module coordination.
    These tasks are particularly challenging for mobile manipulators as they require the simultaneous optimization of base positioning and arm reaching, where minor misalignments in the platform's pose propagate as significant errors at the end-effector.
    }}
    \label{fig:tasks_vis}
\end{figure*}

\textbf{Baselines and Ablation Build-Up.}
We compare against two baselines that share the same residual learning paradigm but differ in supervision: \textit{+HG-DAgger}~\cite{kelly2019hg} learns a residual via MSE on human corrections, and \textit{+Flat Residual} (equivalent to HIL-SERL Residual~\cite{luo2025precise}) is an off-policy RL baseline operating on a monolithic residual vector. To isolate the contribution of each HiRe-MoMa component, we evaluate the following build-up chain atop a frozen base policy:
\textit{(i) +Flat Residual}: single residual head over the joint action space (no hierarchy);
\textit{(ii) +Hierarchical Decoupling}: splits the residual into independent base/arm heads with unidirectional conditioning (\autoref{sec:method_hierarchical_residual_architecture});
\textit{(iii) +Alignment (Full HiRe-MoMa)}: adds the Expert-Guided Value Alignment loss atop hierarchical decoupling (\autoref{sec:method_hil_auxiliary_loss}).
We use Diffusion Policy (DP)~\cite{chi2025diffusion} as the default base policy and additionally evaluate ACT~\cite{Zhao-RSS-23} to verify policy-agnostic generalization, as shown in~\reftab{tab:exp_main_agnostic}.

\textbf{Tasks and Protocol.}
We design three real-world tasks of increasing complexity (\autoref{fig:tasks_vis}):
\textit{Place Only} (single-arm placement),
\textit{Sweep Litter} (dual-arm coordination), and
\textit{Pick Place} (long-horizon base--arm synchronization),
with maximum episode lengths of $500$, $500$, and $800$ steps at 10\,Hz,
beyond which a trial counts as a failure.
\textit{Pick Place} uses a phased success metric ($0.5$ for grasp, $1.0$ for placement);
the others use binary success.
Each task is evaluated over $20$ trials with randomized initialization (identical in training and evaluation),
and we report both \textit{success rate} and \textit{episode length}.
The residual policy and critic are MLPs (hidden $256$) trained with off-policy SAC~\cite{luo2025precise};
the residual is constrained by $\| \mathbf{a}^{\text{res}} \| \le \epsilon = 0.3$,
and the alignment-loss weight is $\lambda_{\text{imi}} = 10^{-3}$.
Per-task action dimensions, demo counts, and remaining hyperparameters are listed in App.~\ref{appendix:exp_setup}--\ref{appendix:impl_details}.

\subsection{Comparative Evaluation: Performance and Generality}
\label{sec:exp_main_results}
As shown in \reftab{tab:exp_main_results}, the full HiRe-MoMa reaches $95\%$ success on \textit{Place Only} and $90\%$ on \textit{Sweep Litter} within $25$--$40$ minutes of real-world fine-tuning, and lifts the long-horizon \textit{Pick Place} from $22.5\%$ (frozen base) to $40\%$ (failure modes in~\refapp{appendix:failure_modes}).
In contrast, \textit{+HG-DAgger} suffers from overfitting and erratic corrections under purely imitation-based supervision, while \textit{+Flat Residual} (HIL-SERL Residual~\cite{luo2025precise}) shows only marginal improvement: its monolithic action space cannot resolve the coupled base-arm dynamics.
HiRe-MoMa also consistently reduces episode length, indicating improved execution efficiency.
\begin{wraptable}{r}{0.58\linewidth}
    \centering
    \caption{\small{
    \textbf{Policy-Agnostic Generalization.}
    Our HiRe-MoMa yields consistent gains when applied to ACT~\cite{Zhao-RSS-23}, demonstrating architecture-independent efficacy.
    While the performance ceiling is inherently tied to the base policy's quality, the results confirm HiRe-MoMa's capacity to fine-tune diverse base policies.
    }}
    \scalebox{0.70}{
    \begin{tabular}{r|c|cc|cc}
    &\multirow{2}{*}{\makecell[c]{\textbf{Training}\\\textbf{Time}\\\textbf{(mins)}}}
    &\multicolumn{2}{c}{\textbf{Success Rate (\%)}}
    &\multicolumn{2}{c}{\textbf{Episode length (steps)}}
    \\
    \textbf{Task} &
    &\makecell[c]{Base\\(ACT)}
    &\makecell[c]{HiRe-MoMa\\(Full)}
    &\makecell[c]{Base\\(ACT)}
    &\makecell[c]{HiRe-MoMa\\(Full)}
    \\
    \hline
    Place Only   &25 &40 &80 (+100\%) &381 &303 (1.26x) \\
    Sweep Litter &30 &40 &75 (+88\%) &362 &297 (1.22x) \\
    \hline
    \rowcolor{table_highlight}
    \textbf{Average}  &27.5 &40.0 &77.5 (+94\%) &371.5 &300.0 (1.24x) \\
    \end{tabular}
    }
    \label{tab:exp_main_agnostic}
\end{wraptable}
To further verify the framework's versatility, we evaluate HiRe-MoMa using ACT~\cite{Zhao-RSS-23} (a Transformer-based policy) as the prior.
Results in \reftab{tab:exp_main_agnostic} show a 94\% relative improvement in success rate.
While the absolute performance is slightly bounded by the base policy's initial coverage, consistent with recent findings~\cite{ankile2025imitation,gong2025robust},
the consistent gains across disparate architectures (ACT and DP)
support \textbf{RQ1}.

\begin{figure}[!t]
    \centering
    \includegraphics[width=\linewidth]{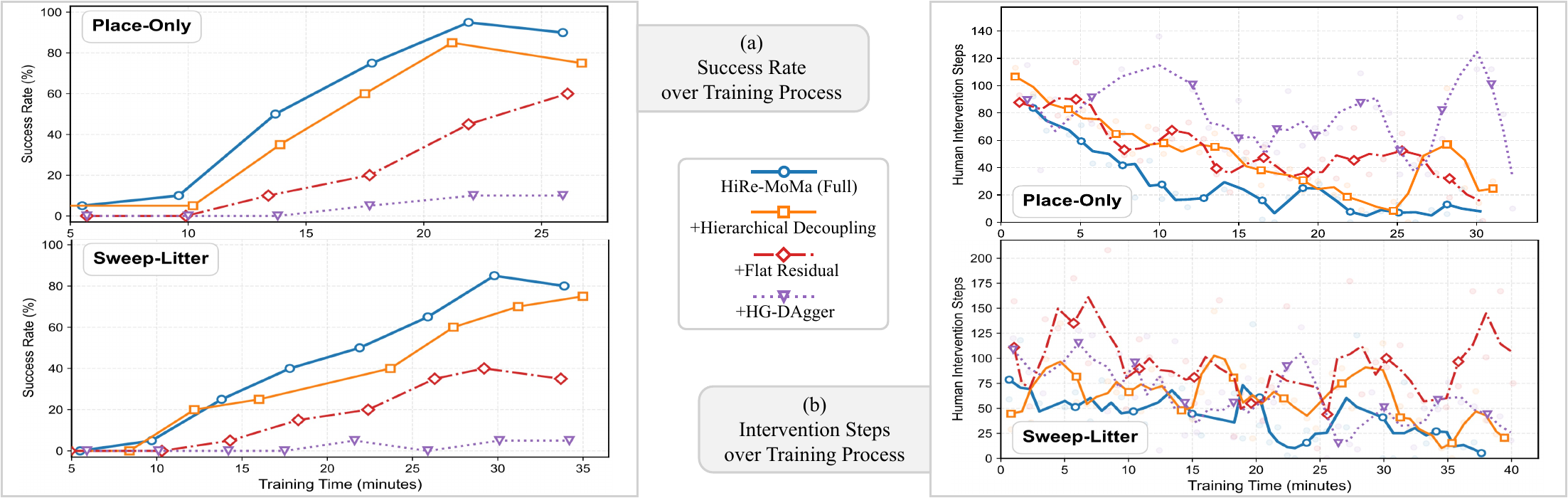}
    \caption{\small{
    \textbf{Training Dynamics.}
    Comparison across methods over training time, where we construct the figure along the actual training time.
    \textbf{(a)} Success rate: \textit{HiRe-MoMa(full)} attains a consistently higher success rate throughout training, validating the effectiveness of our design combining hierarchical residual architecture and alignment.
    \textbf{(b)} Human intervention steps: HiRe-MoMa requires fewer human interventions and converges faster than the baselines, and
    the comparison with and without
    alignment (\textit{+Hierarchical Decoupling}) demonstrates the effectiveness of the extra alignment loss.
    }}
    \label{fig:training_curves}
\end{figure}

\subsection{Build-Up Ablation: Hierarchical Decoupling and Value Alignment}
\label{sec:exp_ablation_studies}
\reftab{tab:exp_ablation_buildup} reports each component's contribution along the build-up chain, and \autoref{fig:training_curves}(a) visualizes the corresponding training-time success-rate curves on \textit{Place Only} and \textit{Sweep Litter}, where the gain from \textit{+Hierarchical Decoupling} is clearly visible (\textcolor{figure_orange}{line}).
\textbf{Effect of Hierarchical Decoupling (RQ2).}
Moving from \textit{+Flat Residual} to \textit{+Hierarchical Decoupling} raises the average success rate from $50.0\%$ to $85.0\%$ and shortens episode length from $361.5$ to $293.5$ steps.
Splitting the residual into base/arm heads with unidirectional conditioning mitigates the gradient interference of a monolithic action space, allowing each sub-policy to converge more effectively,
which is a finding consistent with recent studies on modular visuomotor learning~\cite{gong2025carp}.
The arm head, conditioned on the base correction, generates refinements that are reactive to the platform's displacement, mirroring the system's physical hierarchy and stabilizing coordinated behavior.
\begin{wrapfigure}{r}{0.42\linewidth}
    \centering
    \includegraphics[width=\linewidth]{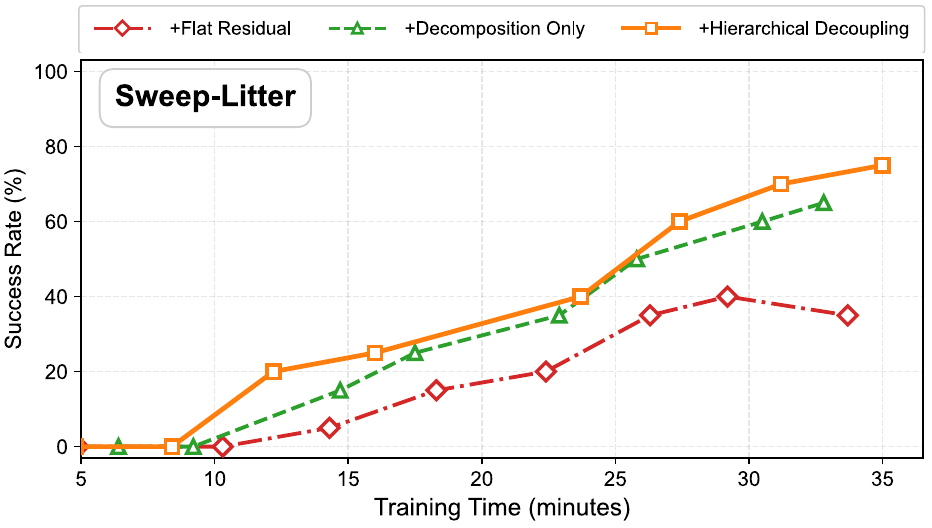}
    \caption{\small{
    \textbf{Disentangling Decomposition and Conditioning.}
    Training-time success-rate curves for \textit{+Flat Residual}, \textit{+Decomposition-Only}, and \textit{+Hierarchical Decoupling} on \textit{Place Only} and \textit{Sweep Litter}; see~\refapp{appendix:split_vs_condition}.
    }}
    \label{fig:decom_or_cond}
\end{wrapfigure}
To further disentangle the contributions of decomposition and unidirectional conditioning within Hierarchical Decoupling, we additionally evaluate a \textit{+Decomposition-Only} variant; \autoref{fig:decom_or_cond} indicates that decomposition contributes the majority of the gain, with conditioning adding a smaller, task-dependent improvement (full breakdown in~\refapp{appendix:split_vs_condition}).
\textbf{Effect of Expert-Guided Value Alignment (RQ3).}
Adding the alignment loss on top of hierarchical decoupling (\textit{+Alignment, Full HiRe-MoMa}) further improves the average success rate to $92.5\%$ and reduces episode length to $253.0$ steps,
while the human-intervention ratio decreases monotonically over training when the alignment loss is added (\autoref{fig:training_curves}(b), \textcolor{figure_blue}{line}; see also~\refapp{appendix:impl_intervention}).
By penalizing the critic for deviating from expert preferences, the value landscape is reshaped to favor concise corrections over redundant RL exploration, enabling the robot to complete tasks with fewer steps and less human supervision.

\subsection{Qualitative Analysis: Interpretability and Visualization}
\label{sec:exp_analysis_visualization}
\begin{figure}[!ht]
   \centering
   \includegraphics[width=\linewidth]{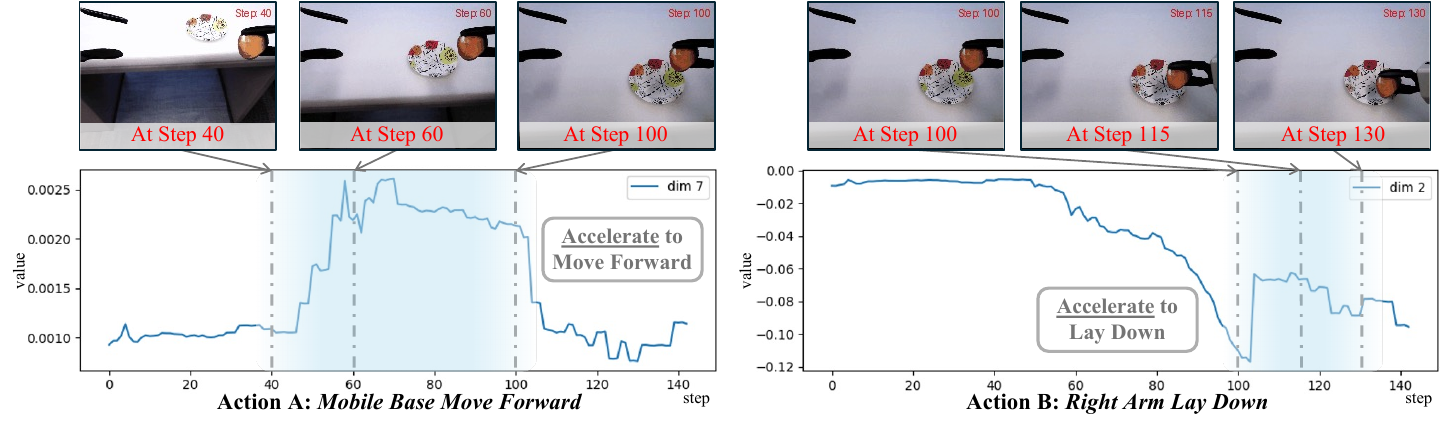}
   \caption{\small{
   \textbf{Residual Correction Visualization.}
   Base longitudinal velocity and manipulator vertical position throughout the \textit{Place Only} task.
   HiRe-MoMa applies coherent corrections at critical trajectory phases, most notably during base acceleration and the manipulator's lay-down motion.
   }}
   \label{fig:analysis_correction_visualization}
\end{figure}

To probe what the residual policy actually learns, we visualize per-dimension corrections on the \textit{Place Only} task (\autoref{fig:analysis_correction_visualization}), focusing on the base's forward velocity and the arm's downward placement motion.
The corrections concentrate at critical decision boundaries: even bounded by $\| \mathbf{a}^{\text{res}} \| \le \epsilon$, they compensate for the base's inertial drift and the arm's kinematic inaccuracies, refining the base policy's coarse trajectory into more precise control signals.
The pattern indicates that the learned correction policy is physically interpretable and consistent with the platform's hierarchy
(\textbf{RQ4}).

\section{Limitations and Future Work}
\label{sec:limitation_futurework}
Our evaluation covers a single platform (TIAGo++) with $20$ trials per condition and no seed sweeps, and \textit{Pick Place} remains bounded by platform hardware.
The two residual heads share one sparse reward without explicit hierarchical credit assignment~\cite{levy2019learning, nachum2018data},
so arm failures and human-assisted successes may bias the base critic;
training also needs a small warm-up buffer of successful on-robot rollouts.
Finally, each run requires a supervising operator,
and the policies are not yet reliable for unsupervised deployment around people.
Future work includes contact-rich tasks with multimodal sensing, richer inter-module structures, and validation on more platforms.
See~\refapp{appendix:limitations} for details.

\section{Conclusion}
\label{sec:conclusion}
We presented HiRe-MoMa, a real-world residual fine-tuning framework for mobile manipulation that combines a policy-agnostic residual, hierarchical decoupling with unidirectional conditioning, and an expert-guided value-alignment loss.
Across three TIAGo++ tasks, HiRe-MoMa raises average success from $39.2\%$ to $75\%$ within $25$--$40$\,min of online interaction, shortens execution by $1.29\times$, and remains effective across base architectures with less than $20\%$ human intervention.

\bibliography{references}
\clearpage
\appendix
\textbf{\Large{Appendix}}
\vspace{5pt}

\section{Additional Experiments}
\label{appendix:extended_experiments}

\subsection{Disentangling Decomposition and Conditioning}
\label{appendix:split_vs_condition}

The \textit{+Hierarchical Decoupling} column in \reftab{tab:exp_ablation_buildup} combines two sub-operations introduced in \autoref{sec:method_hierarchical_residual_architecture}.
\textit{(i) decomposition}: splitting the residual into independent base/arm heads ($\mathbf{a}^{\text{res}} = [\mathbf{a}_{\text{mo}}^{\text{res}}, \mathbf{a}_{\text{ma}}^{\text{res}}]$);
and
\textit{(ii) unidirectional conditioning}: feeding the base correction into the arm head (\refeq{eq:hierarchy_condition}).
To isolate their individual contributions, we additionally evaluate a \textit{+Decomposition-Only} variant in which the base and arm heads are trained independently without inter-module information flow. \reftab{tab:split_vs_condition} reports the results.

\begin{table}[ht]
    \centering
    \caption{\small{
    \textbf{Decomposition vs.\ Conditioning within Hierarchical Decoupling.}
    Decomposition alone provides the dominant gain by mitigating gradient interference ($+30$\,pt average success rate).
    Conditioning adds a targeted improvement ($+5$\,pt on average) that manifests on the coupled dual-arm task (\textit{Sweep Litter}).
    }}
    \scalebox{0.80}{
    \begin{tabular}{r|ccc|ccc}
    &\multicolumn{3}{c}{\textbf{Success Rate (\%)}}
    &\multicolumn{3}{c}{\textbf{Episode Length (steps)}}
    \\
    \textbf{Task}
    &\makecell[c]{+Flat\\Residual}
    &\makecell[c]{+Decomposition\\Only}
    &\makecell[c]{+Hierarchical\\Decoupling}
    &\makecell[c]{+Flat\\Residual}
    &\makecell[c]{+Decomposition\\Only}
    &\makecell[c]{+Hierarchical\\Decoupling}
    \\
    \hline
    Place Only   &50$\rightarrow$60  &50$\rightarrow$85  &50$\rightarrow$85  &342$\rightarrow$329 &342$\rightarrow$291 &342$\rightarrow$283 \\
    Sweep Litter &45$\rightarrow$40  &45$\rightarrow$75  &45$\rightarrow$85  &359$\rightarrow$394 &359$\rightarrow$322 &359$\rightarrow$304 \\
    \hline
    \rowcolor{table_highlight}
    \textbf{Average} &47.5$\rightarrow$50.0 &47.5$\rightarrow$80.0 &47.5$\rightarrow$85.0 &350.5$\rightarrow$361.5 &350.5$\rightarrow$306.5 &350.5$\rightarrow$293.5 \\
    \end{tabular}
    }
    \label{tab:split_vs_condition}
\end{table}

Decomposition raises the average success rate from $50.0\%$ to $80.0\%$ ($+30$\,pt) and reduces average episode length from $361.5$ to $306.5$ steps, confirming that the primary benefit stems from eliminating gradient interference between heterogeneous modules, in line with prior observations in modular visuomotor learning~\cite{gong2025carp}.
Adding unidirectional conditioning yields a further gain to $85.0\%$ ($+5$\,pt) with $13$ fewer steps on average.
Notably, the conditioning gain is \emph{task-dependent}: negligible on \textit{Place Only} (single-arm, $85\% \to 85\%$) but pronounced on \textit{Sweep Litter} (dual-arm, $75\% \to 85\%$), where base displacement simultaneously affects both arms' reachability.
This indicates that conditioning provides targeted improvements that scale with inter-module coupling complexity, complementary to the broader stabilization achieved by decomposition alone.
We therefore view conditioning as a coupling-dependent refinement on top of decomposition rather than the sole source of the gain; both are components of Hierarchical Decoupling.

\subsection{Training Steps vs.\ Wall-Clock Time}
\label{appendix:training_steps_2_time}

\begin{figure}[!ht]
   \centering
   \includegraphics[width=0.92\linewidth]{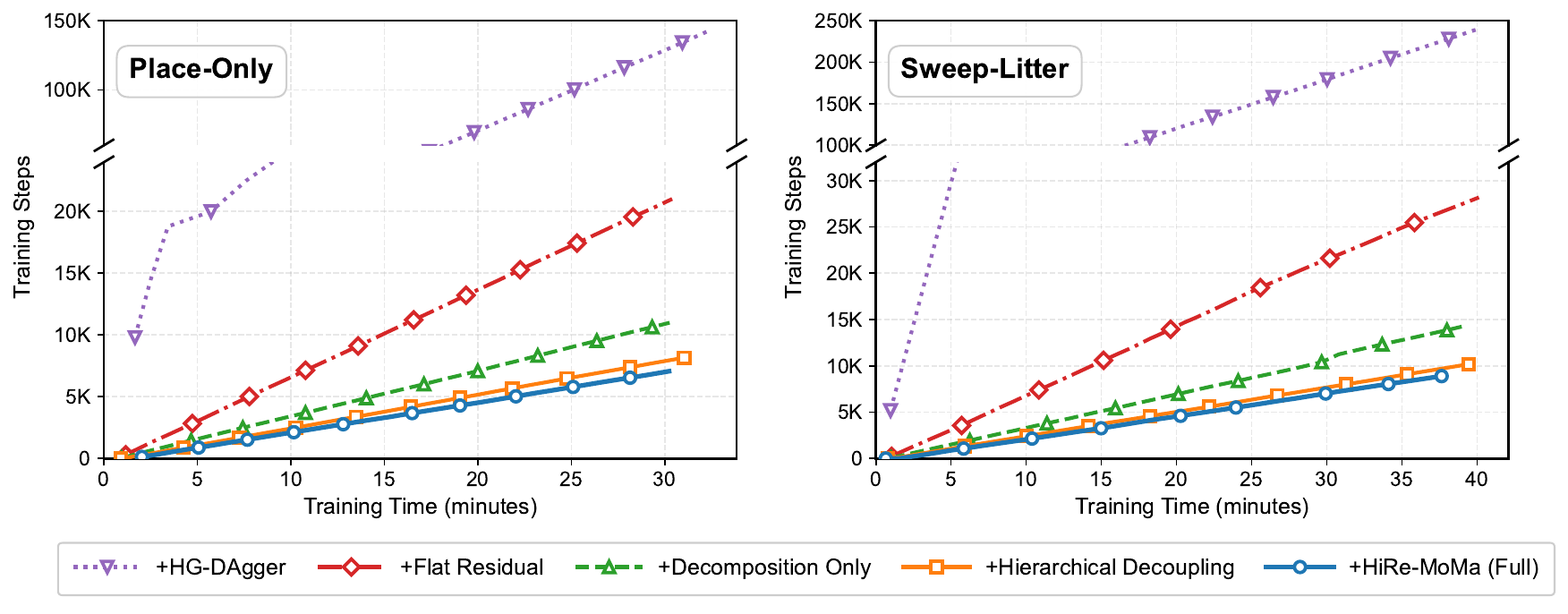}
   \caption{\small{
   \textbf{Training-Step Progression vs.\ Wall-Clock Time.}
   HG-DAgger advances through training steps faster than the SAC-based residual methods, but its higher step throughput does not translate into higher success rates (\autoref{fig:training_curves}(a)). HiRe-MoMa adds only a modest per-step overhead over \textit{+Flat Residual} ($1$--$2\times$) while delivering clear gains in success rate and episode length.
   }}
   \label{fig:training_steps_2_time}
\end{figure}

To enable a fair comparison of training efficiency, we align all methods on a common wall-clock budget and report the corresponding training-step progression in~\autoref{fig:training_steps_2_time}.
HG-DAgger advances through training steps faster owing to its supervised objective, but this does not translate into higher success rates (\autoref{fig:training_curves}(a)), reflecting overfitting to sparse human corrections.
Among the RL-based methods, HiRe-MoMa incurs only a modest overhead relative to \textit{+Flat Residual}~\cite{luo2025precise}, roughly $1$--$2\times$ the per-step compute, since the residual is split into separate base and arm policies, an acceptable cost given the consistent gains in success rate and episode length reported in~\reftab{tab:exp_main_results}.
The wall-clock gap in~\autoref{fig:training_steps_2_time} thus has two additive sources: each SAC update back-propagates through a critic ensemble with min-ensemble TD targets and separate actor and temperature updates, making \textit{+Flat Residual} roughly an order of magnitude slower per step than HG-DAgger's single regression step, and HiRe-MoMa's extra $1$--$2\times$ comes from evaluating its base and arm heads sequentially.

\section{Pseudo Code}
\label{appendix:pseudocode}

We provide pseudocode for the full system: \refalg{algo:hire_moma} details the SAC-style training loop with the
\textit{Expert-Guided Value Alignment} loss, and \refalg{algo:hire_moma_hierarchy} specifies the \textit{Hierarchical Decoupling} design. The corresponding method derivation appears in~\autoref{sec:method}.

\begin{algorithm}[H]
\caption{HiRe-MoMa: SAC-Style Residual Policy w/ HIL }
\label{algo:hire_moma}
\begin{algorithmic}[1]
\State \textbf{Input}: Pre-trained base policy $\pi^{\text{base}}$, warm-up buffer $\mathcal{D}_{\text{warm}}$ of successful on-robot rollouts, human expert $\pi^{\text{res,h}}$.
\State \textbf{Initialize}: Residual policy $\pi^{\text{res}}_{\theta}$, Ensemble critics $Q_{\phi_{1 \dots N}}$, Temperature $\alpha$, Target entropy $\bar{\mathcal{H}}$. \textcolor{algo_comment}{\Comment{See \refalg{algo:hire_moma_hierarchy} for the module-level expansion.}}
\State \textbf{Initialize}: Replay buffer $\mathcal{D}_{\text{replay}} \leftarrow \emptyset$, and expert buffer $\mathcal{D}_{\text{expert}} \leftarrow \mathcal{D}_{\text{warm}}$. \textcolor{algo_comment}{\Comment{Not $\pi^{\text{base}}$'s training data; see~\refapp{appendix:impl_basepolicy}.}}
\State \textbf{Initialize}: Target networks $\phi'_i \leftarrow \phi_i$.
\Statex \textbf{\# --- Rollout Worker ---}
    \State Reset environment and receive initial observation $\mathbf{o}_t$.
    \For{each step $t$}
        \State Compute base action $\mathbf{a}^{\text{base}}_t \sim \pi^{\text{base}}(\mathbf{o}_t)$.
        \If{Human intervenes}
            \State Obtain residual $\mathbf{a}^{\text{res}}_t = \pi^{\text{res,h}}(\mathbf{o}_t, \mathbf{a}^{\text{base}}_t)$.
            \State Flag $\mathbb{I}_{\text{h}} = 1$.
        \Else
            \State Sample residual $\mathbf{a}^{\text{res}}_t \sim \pi^{\text{res}}_{\theta}(\mathbf{o}_t, \mathbf{a}^{\text{base}}_t)$.
            \State Flag $\mathbb{I}_{\text{h}} = 0$.
        \EndIf
        \State Execute $\mathbf{a}^{\text{exe}}_t = \mathbf{a}^{\text{base}}_t + \mathbf{a}^{\text{res}}_t$, observe $(r_t, \mathbf{o}_{t+1}, d_t)$.
        \State Store $(\mathbf{o}_t, \mathbf{a}^{\text{base}}_t, \mathbf{a}^{\text{exe}}_t, r_t, d_t, \mathbf{o}_{t+1}, \mathbf{a}^{\text{base}}_{t+1})$ in $\mathcal{D}_{\text{replay}}$.
        \State \textbf{if} $\mathbb{I}_{\text{h}} = 1$ \textbf{then} add to $\mathcal{D}_{\text{expert}}$.
    \EndFor
\Statex \textbf{\# --- Central Learner ---}
    \For{each update step}
        \State Evenly sample $\mathcal{B}_{\text{replay}} \sim \mathcal{D}_{\text{replay}}$ and $\mathcal{B}_{\text{expert}} \sim \mathcal{D}_{\text{expert}}$.
        \State \# --- Actor Update (using $\mathcal{B} = \mathcal{B}_{\text{replay}} \cup \mathcal{B}_{\text{expert}}$) ---
\State Optimize $\theta$ through $[ \frac{1}{N}\sum_{i}^N Q_{\phi_i}(\mathbf{o}_t, \pi^{\text{res}}_{\theta}(\mathbf{o}_t, \mathbf{a}^{\text{base}}_t) + \mathbf{a}^{\text{base}}_t) - \alpha \log \pi^{\text{res}}_{\theta}(\mathbf{a}_t^{\text{res}} | \mathbf{o}_t) ]$.
        \State Optimize $\alpha$ through $[-\alpha (\log \pi_{\theta}^{\text{res}}(\mathbf{a}_t^{\text{res}} | \mathbf{o}_t) + \bar{\mathcal{H}})]$
        \State \# --- Critic Update (using $\mathcal{B}$) ---
        \State $\mathbf{a}^{\text{exe}}_{t+1} \leftarrow \mathbf{a}^{\text{base}}_{t+1} + \mathbf{a}^{\text{res}}_{t+1} = \mathbf{a}^{\text{base}}_{t+1} + \pi^{\text{res}}_{\theta}(\mathbf{o}_{t+1}, \mathbf{a}^{\text{base}}_{t+1})$
        \State Compute $y = r_t + \gamma (1-d_t) [ \min_{i} Q_{\phi'_i}(\mathbf{o}_{t+1},\mathbf{a}^{\text{exe}}_{t+1})]$.
        \State Update $\phi_i$ by minimizing $\sum_i^N \| Q_{\phi_i}(\mathbf{o}_t, \mathbf{a}^{\text{exe}}_t) - y \|^2$.
\State \textcolor{algo_highlight}{\# --- Expert-Guided Critic Alignment (applied to $\mathcal{B}_{\text{expert}}$ only) ---}
        \State \textcolor{algo_highlight}{Additionally optimize $\phi_i$ through
        $\sum_i \log (1 + \exp(Q_{\phi_i}$ $(\mathbf{o}_t, \pi^{\text{res}}_{\theta}(\mathbf{o}_t,\mathbf{a}_t^{\text{base}}) + \mathbf{a}^{\text{base}}_t) - Q_{\phi_i}(\mathbf{o}_t, \mathbf{a}^{\text{exe}}_t)))$.
        }
        \textcolor{algo_comment}{\Comment{See \autoref{sec:method_hil_auxiliary_loss} for details.}}
        \State Update target critic networks: $\phi'_i \leftarrow \rho \phi'_i + (1-\rho) \phi_i$.
        \State Synchronize latest policy $\theta$ to rollout worker.
    \EndFor
\end{algorithmic}
\end{algorithm}

\begin{algorithm}[H]
\caption{HiRe-MoMa: Hierarchical Decoupling Design}
\label{algo:hire_moma_hierarchy}
\begin{algorithmic}[1]
\State \textbf{Initialize}: Separate residual policies $\pi^{\text{res}}_{\theta_{\text{mo}}}$, $\pi^{\text{res}}_{\theta_{\text{ma}}}$ and critics $Q_{\phi_{\text{mo}}}$, $Q_{\phi_{\text{ma}}}$.
\textcolor{algo_comment}{\Comment{Critic ensembles omitted for brevity.}}
\State \textbf{Notation}: $\mathbf{a}_t = [\mathbf{a}_{t, \text{mo}}, \mathbf{a}_{t, \text{ma}}]$.
\Statex \# --- Hierarchical Action Generation ---
    \State Mobile Base Residual: $\mathbf{a}_{t, \text{mo}}^{\text{res}} \sim \pi^{\text{res}}_{\theta_{\text{mo}}}(\mathbf{o}_t, \mathbf{a}_{t, \text{mo}}^{\text{base}})$.
    \State \textcolor{algo_highlight}{Manipulator Arm Residual: $\mathbf{a}_{t, \text{ma}}^{\text{res}} \sim \pi^{\text{res}}_{\theta_{\text{ma}}}(\mathbf{a}_{t, \text{mo}}^{\text{res}}, \mathbf{o}_t, \mathbf{a}_{t, \text{ma}}^{\text{base}})$.}
    \State \textcolor{algo_highlight}{Manipulator Arm Critic: $Q_{\phi_{\text{ma}}}(\mathbf{a}_{t, \text{mo}}^{\text{res}}, \mathbf{o}_t, \mathbf{a}_{t, \text{ma}}^{\text{exe}})$.}
    \State Combined Executed Action: $\mathbf{a}_t^{\text{exe}} = \mathbf{a}_t^{\text{base}} + [\mathbf{a}_{t, \text{mo}}^{\text{res}}, \mathbf{a}_{t, \text{ma}}^{\text{res}}]$.
\Statex \# --- Hierarchical Policy Optimization ---
    \State Update Base and Arm modules independently with unidirectional flow. \textcolor{algo_comment}{\Comment{Decouples base-arm gradient interference during fine-tuning.}}
    \State Update $\theta_{\text{mo}}$ via $Q_{\phi_{\text{mo}}}$ using standard gradient.
    \State Update $\theta_{\text{ma}}$ via $Q_{\phi_{\text{ma}}}$ while conditioning on $\mathbf{a}_{t, \text{mo}}^{\text{res}}$.
\end{algorithmic}
\end{algorithm}

\section{Experimental Setup}
\label{appendix:exp_setup}

\subsection{Hardware and Control Interface}
\label{appendix:exp_hardware}
We evaluate HiRe-MoMa on the TIAGo++ mobile manipulator, which features two 7-DoF arms, a liftable torso, a wheeled mobile base, and a head equipped with an RGB camera.
To manage the complexity of real-world RL, we fix the torso and head poses, leaving a 17-dimensional action space: a 3-D base velocity and, per arm, a 6-D end-effector delta pose plus a 1-D discrete gripper command, of which each task uses a subset (\refapp{appendix:exp_tasks}).
Base and arm commands are output jointly at every step, so the base and arms move simultaneously rather than in alternating phases.
Visual feedback is provided by a head-mounted RGB camera at $208\times160$ resolution.
For human-in-the-loop (HIL) integration, we employ a teleoperation system~\cite{dass2024telemoma} based on the Meta Quest 3.
The expert simultaneously provides velocity commands for the base and end-effector offsets in Euler space, which are then resolved into joint-space deltas via Inverse Kinematics (IK).
During shared-control rollouts, human interventions preempt the residual policy with absolute priority; arbitration runs locally on the robot's onboard unit (\autoref{fig:teaser}) for low-latency response.

\subsection{Distributed Asynchronous Training}
\label{appendix:exp_distributed}
We adopt a distributed asynchronous training paradigm inspired by HIL-SERL~\cite{luo2025precise}, decomposed into three loosely coupled components:
\textbf{Rollout Worker}: an asynchronous actor process on the robot (not a separate robot station) that executes actions and collects transitions at $10$~Hz; all experiments use a single TIAGo++, although the shared replay buffer would also admit additional robots.
\textbf{Central Learner}: performs continuous policy and critic optimization on a workstation (NVIDIA RTX 4090) from the shared replay buffer, asynchronously to data collection; the rollout worker syncs the latest actor checkpoint every $50$ environment steps.
\textbf{Inference Node}: a standalone service that continuously provides frozen base-policy predictions as conditioning observations for the residual agent.
Communication between the onboard unit and the workstation runs over TCP/IP, providing decoupled synchronization with low end-to-end latency.

\subsection{Tasks and Evaluation Protocol}
\label{appendix:exp_tasks}
We evaluate on three real-world tasks of increasing complexity (\autoref{fig:tasks_vis}).
In each trial we randomize the robot's initial pose ($\pm 30^\circ$ orientation, $1.5$\,m starting distance from the table where applicable) and manually perturb object placements (bread, plate, litter) by roughly $10$--$20$\,cm around nominal positions, using the same randomization for training and evaluation:
\textbf{(I) \textit{Place Only} (Single-Arm).} The robot starts holding bread in its right gripper and must navigate to the table and precisely place the bread onto a plate. Action space: $\mathcal{A} \in \mathbb{R}^{10}$ (6-D right-arm end-effector delta + 1-D gripper + 3-D base); maximum episode length: $500$ steps ($50$\,s at $10$\,Hz).
\textbf{(II) \textit{Sweep Litter} (Dual-Arm).} TIAGo++ coordinates a broom (right arm) and a dustpan (left arm) to sweep litter into the bag. Action space: $\mathcal{A} \in \mathbb{R}^{15}$ (two 6-D end-effector deltas + 3-D base; the two gripper dimensions are dropped because both hands hold their tools throughout); maximum episode length: $500$ steps. This task tests high-dimensional bilateral coordination.
\textbf{(III) \textit{Pick Place} (Long-Horizon).} Starting empty-handed at $1.5$\,m from the source table, the robot must (1) approach and grasp the bread, (2) transition between two tables, and (3) place it on a target plate. Action space: $\mathcal{A} \in \mathbb{R}^{10}$ (6-D end-effector delta + 1-D gripper + 3-D base); maximum episode length: $800$ steps ($80$\,s). This is the most demanding task, requiring tight base--arm spatiotemporal synchronization (\autoref{fig:teaser}).

For each task we run $20$ independent trials with randomized initialization. A trial is marked as failure if it exceeds the maximum step limit or violates task-specific safety constraints. Success is task-specific:
\textbf{\textit{Place Only}}: the bread must be released and stabilized within the target plate.
\textbf{\textit{Sweep Litter}}: all target litter must be fully swept into the dustpan bag; any litter remaining outside or displaced beyond the workspace is a failure.
\textbf{\textit{Pick Place}}: we use a phased success metric---successful grasping and lifting yield a partial score of $0.5$, and successful placement on the target plate yields the full score $1.0$.
For all tasks we report \textit{success rate} (\%) and \textit{episode length} (steps to completion) averaged over the $20$ trials.

\subsection{Failure Modes and Pick Place Ablation}
\label{appendix:exp_pickplace}
\label{appendix:failure_modes}
Before fine-tuning, the frozen DP fails $10/20$ trials on \textit{Place Only} ($6$ base mis-positioning, $4$ wrong release height) and $11/20$ on \textit{Sweep Litter} ($7$ broom--dustpan misalignment from base drift, $4$ incomplete sweeps), i.e., mostly base-positioning errors that propagate to the arm; HiRe-MoMa reduces these to $1/20$ and $2/20$.
On \textit{Pick Place}, the remaining failures stem mainly from hardware: the arm's narrow workspace turns small base errors after the table-to-table transition into IK-infeasible grasp targets, and the casters stick after prolonged operation, both beyond what bounded residuals ($\epsilon = 0.3$) can absorb.
Because the long table-to-table transition aggravates these hardware issues and the $800$-step episodes are costly to repeat across ablated variants, \textit{Pick Place} is reported only in the main comparison (\reftab{tab:exp_main_results}) and omitted from the build-up ablation (\reftab{tab:exp_ablation_buildup}).

\section{Implementation Details}
\label{appendix:impl_details}

\subsection{Observations and Actions}
\label{appendix:impl_obs}
The observation space consists of egocentric RGB images ($208\times160$) and a $\sim 20$-dimensional proprioceptive vector (whose exact dimension is task-dependent and kept consistent across all baselines). We use a relative action space, with all continuous dimensions normalized to $[-1, 1]$. The gripper is controlled discretely; all other dimensions are continuous. The residual correction magnitude is bounded by $\| \mathbf{a}^{\text{res}} \| \le \epsilon$ with $\epsilon = 0.3$, as introduced in \autoref{sec:method_hierarchical_residual_architecture}.

\subsection{Network Architecture and Optimization}
\label{appendix:impl_network}
Both actor and critic of the residual policy are parameterized by MLPs with hidden dimension $256$. We use an off-policy SAC framework~\cite{luo2025precise} with a critic ensemble for conservative target estimation (\refalg{algo:hire_moma}). The auxiliary alignment-loss weight is fixed at $\lambda_{\text{imi}} = 10^{-3}$ across all tasks; target critic networks are updated via Polyak averaging following standard SAC. The base policy remains \emph{frozen} throughout fine-tuning to preserve the pre-trained priors.
The residual bound $\epsilon$ trades correction capacity against exploration stability: too small an $\epsilon$ under-corrects the base policy, while too large an $\epsilon$ enlarges the exploration space and destabilizes SAC, consistent with prior residual RL~\cite{jiang2025transic, ankile2025imitation}; $\epsilon = 0.3$ works well on all three tasks.

\subsection{Base-Policy Pre-Training and Warm-Up Buffer}
\label{appendix:impl_basepolicy}
We pre-train two frozen base policies, Diffusion Policy (DP)~\cite{chi2025diffusion} and ACT~\cite{Zhao-RSS-23}, on human demonstrations: $80$ trajectories for \textit{Place Only}, $100$ for \textit{Sweep Litter}, and $200$ for \textit{Pick Place}. Each base model is trained for $1{,}000$ epochs under the standard configuration of its respective architecture. For DP we set observation horizon $1$, prediction horizon $8$, and action horizon $6$, following standard visuomotor practice~\cite{gong2025carp}; ACT is configured with matching chunk sizes.
These demonstrations are used only to pre-train $\pi^{\text{base}}$.
For fine-tuning, following HIL-SERL~\cite{luo2025precise}, we instead warm-start the expert buffer with $\sim$500--600 transitions from successful episodes collected on the robot by rolling out the frozen base policy, with operator takeover when needed ($\mathcal{D}_{\text{warm}}$ in~\refalg{algo:hire_moma}); HiRe-MoMa thus never accesses the base policy's training data.

\subsection{Human Intervention Protocol and Budget}
\label{appendix:impl_intervention}
A single trained operator supervises all training runs, blind to the method being trained, and intervenes when failure is imminent but still recoverable (e.g., the gripper starts releasing while misaligned with the plate, or the broom is extended well wide of the litter).
Interventions are collected online within each method's own run under this identical protocol, so all methods receive the same intervention opportunities.
We measure the budget by the \textit{intervention coverage}, i.e., the fraction of human-intervened transitions among all collected transitions: it stays below $20$\% on all tasks and decreases from roughly $15$--$20$\% early in training to $0$--$3$\% near convergence, as interventions concentrate at task-critical phases (\autoref{fig:training_curves}(b)).
Since interventions reshape the critic rather than being cloned, the method tolerates moderate teleoperation noise; a systematic study of operator expertise is left to future work.

\section{Related Work}
\label{appendix:related_work}

\subsection{Real-World Reinforcement Learning}
\label{appendix:rw_rwrl}
Real-world RL aims to bridge the sim-to-real gap through direct physical interaction~\cite{XuC1-RSS-25}. Among the persistent bottlenecks, three lines of work are particularly relevant.
\textbf{(i) Reward specification.} To reduce the burden of reward engineering, recent studies have leveraged vision-language representations~\cite{Ma2023LIVLR}, video-based goal rewards~\cite{yu2025genflowrl}, and language-conditioned prediction models~\cite{zhang2025rewind} as alternatives to hand-crafted dense rewards. HiRe-MoMa instead adopts a sparse binary success reward, removing the need for these auxiliary signals.
\textbf{(ii) Autonomous learning loops.} Frameworks such as MEDAL~\cite{sharma2022state, sharma2023self} and other self-resetting systems~\cite{pmlr-v305-zhou25a} use forward-backward policy pairs to automate task resets. While complementary, these techniques are orthogonal to our focus on residual fine-tuning.
\textbf{(iii) Sample efficiency.} Early on-policy deployments~\cite{zhu2019dexterous} demonstrated feasibility, after which the field moved toward model-based~\cite{wu2023daydreamer} and off-policy~\cite{hu2023reboot} architectures. More recent advances include Mixture-of-Experts gradient deconfliction~\cite{huang2025mentor}, simulation-based pre-training~\cite{wu2025simlauncher}, and warm-start RL for offline-to-online transitions~\cite{zhou2025efficient}. The SERL framework~\cite{luo2024serl} consolidated many of these ideas into a practical off-policy paradigm that has inspired follow-on residual fine-tuning systems~\cite{chen2025conrft, ankile2025residual}. HiRe-MoMa builds on this off-policy foundation, but extends it to the unique coupling structure of mobile manipulation,
where prior approaches based on simulation-pre-trained latent action spaces~\cite{hu2025slac} are computationally expensive and lack
adaptability.

\subsection{Policy Post-Training and Residual RL}
\label{appendix:rw_posttraining}
Post-training has emerged as a key paradigm for refining base policies derived from large-scale Vision-Language-Action models~\cite{zhao2025vlas} or specialized imitation-learning frameworks~\cite{chi2025diffusion}. Interactive imitation learning~\cite{kelly2019hg, hu2025rac} incorporates human supervision to mitigate distributional drift, while RL extends agent capability beyond the offline dataset through autonomous exploration. However, generative base policies pose a technical hurdle: their intractable log-likelihoods preclude standard policy gradient~\cite{schulman2017proximal} or value-based~\cite{haarnoja2018soft} updates. To address this, recent work has formulated surrogate objectives~\cite{zhang2025reinflow, chen2025pi_} or alternative entropy measures~\cite{wang2024diffusion, pmlr-v267-ma25d} to bypass explicit likelihood estimation; nonetheless, these approaches typically require architecture-specific modifications and limit cross-platform portability. Policy-agnostic RL~\cite{mark2024policy} offers a more general route but often suffers from training instability and over-conservatism~\cite{chen2025conrft}.
Residual RL offers a practical alternative by learning additive corrections atop frozen base policies~\cite{ankile2025residual, gong2025robust}; HiRe-MoMa builds on this paradigm and introduces a structured decoupling tailored to MoMa's cross-module coupling.

\subsection{Mobile Manipulation}
\label{appendix:rw_moma}
Mobile manipulation has been studied along several complementary axes. Recent approaches have leveraged Vision-Language Models~\cite{gao2026guides, zhang2026commcp} for high-level reasoning~\cite{shah2025bumble} or large-scale human video datasets~\cite{bahety2025safemimic} to scale base capabilities, while imitation-learning studies show that exploiting hierarchical relationships within modular action spaces is critical for downstream performance~\cite{jiang2025brs, chen2025acdit}.
In the RL domain, however, this structural insight remains under-explored: most methods rely on extensive simulation and hand-crafted constraints, such as restricted latent action spaces, to manage the sim-to-real gap~\cite{Hu-RSS-23, zhang2025catch}.
Even when real-world RL is used for continuous adaptation~\cite{Mendonca2024Continuously}, the vast search space often necessitates task-specific priors that limit cross-scenario generalization, or computationally expensive skill-discovery phases~\cite{hu2025slac}.
In contrast, HiRe-MoMa operates directly in the raw action space of any pre-trained policy and avoids manually tuned reward shaping~\cite{Hu-RSS-23} by exploiting the system's hierarchical dependencies under a sparse, success-only reward.

\subsection{Human-in-the-Loop Robot Learning}
\label{appendix:rw_hil}
Human input~\cite{kim2025human, gong2026autofocus} has long served as a supervisory signal for aligning agent behavior with expert intent~\cite{wang2025inference, korkmaz2025mile}. In the RL setting, interventions are traditionally used as safety buffers to prevent policy collapse during exploration~\cite{luo2025precise, amin2025pi}, treated as data sources for imitation~\cite{Liu-RSS-23}, or used to construct auxiliary losses for denser supervision~\cite{li2022efficient}. Beyond these conventional roles, expert corrections also encode implicit information about the relative value of state-action pairs. HiRe-MoMa exploits this perspective by treating corrections as functional constraints on the critic's value surface, complementing rather than replacing the standard imitation pipeline.

\section{Limitations and Future Work}
\label{appendix:limitations}
\label{appendix:future_work}
\textbf{Limitations.}
Beyond the single-platform scope (TIAGo++, $20$ trials per condition, no seed sweeps, build-up ablation on two of three tasks) and the hardware-bound \textit{Pick Place} results (\refapp{appendix:failure_modes}), HiRe-MoMa has three methodological limitations.
First, its hierarchy is spatial rather than temporal: unlike hierarchical RL methods that assign credit across temporal levels via hindsight relabeling~\cite{levy2019learning} or off-policy subgoal correction~\cite{nachum2018data}, both heads act at the same timescale and share one sparse reward.
Since a base pose is only good relative to the arm's ability to act from it, we model this coupling implicitly by conditioning the arm on the base residual instead of assigning per-module credit.
As a result, an arm failure also lowers the value of a correct base motion, and human-assisted successes can inflate $Q_{\phi_{\text{mo}}}$ at base poses where the autonomous arm would fail; sparse interventions limit but do not remove this bias, since balanced sampling keeps them in every batch.
Second, although HiRe-MoMa never accesses the base policy's training data, it needs a small warm-up buffer of successful on-robot rollouts per task (\refapp{appendix:impl_basepolicy}).
Third, an MLP residual conditioned on the current observation may miss history-dependent corrections of strongly non-Markovian base policies (e.g., action chunking); a history buffer or recurrent head is a natural extension.

While HiRe-MoMa is effective on the evaluated tasks, several directions remain open for future:

\textbf{Adaptive Regularization.}
The residual bound $\epsilon = 0.3$ and the expert-guided alignment weight $\lambda_{\text{imi}} = 10^{-3}$ are currently fixed across tasks, which suffices for our evaluated settings but is unlikely to be optimal across broader task distributions.
Developing adaptive weighting mechanisms,
such as learnable coefficients, curriculum-based scheduling, or human-in-the-loop prior scheduling,
could dynamically adjust the regularization strength based on the agent's real-time performance and task complexity,
potentially improving robustness across diverse task distributions.

\textbf{Multimodal Sensing Integration.}
Our framework primarily utilizes visual and proprioceptive feedback. However, many fine-grained manipulation tasks are inherently contact-rich, requiring precise force and tactile sensing to ensure safety and dexterity. Integrating multimodal fusion (e.g., visuo-tactile or force-torque sensors) into the residual learning process is a critical next step for scaling to complex, industrial-grade applications.

\textbf{Richer Inter-Module Structures.}
HiRe-MoMa's hierarchical decoupling currently uses a two-level structure (base $\to$ arm), which fits the evaluated tasks but does not directly extend to bi-manual coordination with active torso and head control, or to multi-stage long-horizon tasks with temporal dependencies. Such configurations may benefit from deeper or non-tree-shaped inter-module structures; exploring latent representations or graph-based conditioning mechanisms could capture richer relational dependencies.
Larger workspaces with obstacles would further call for collision-aware base corrections, while multi-stage semantic reasoning is best delegated to a stronger base policy (e.g., a VLA~\cite{zhao2025vlas}), with HiRe-MoMa providing the low-level whole-body corrections.

\textbf{Cross-Platform and Simulation Validation.}
To enhance reproducibility and broaden adoption, replicating the HiRe-MoMa protocol in high-fidelity simulators (e.g., Robocasa~\cite{nasiriany2024robocasa}, Isaac Sim, Behavior-1K~\cite{li2023behavior}) and validating on additional mobile manipulation platforms (e.g., Stretch, Fetch) would clarify the framework's generalizability and surface platform-specific tuning requirements.
A platform with a wider arm workspace and more reliable base actuation would also lift the hardware ceiling observed on \textit{Pick Place} (\refapp{appendix:failure_modes}).

\end{document}